\documentclass[10pt,twocolumn,letterpaper]{article}

\usepackage{cvpr}      
\usepackage{xspace}
\usepackage{makecell}
\usepackage{multirow}
\usepackage{caption}
\usepackage[dvipsnames]{xcolor}

\usepackage[symbol]{footmisc}
\definecolor{cvprblue}{rgb}{0.21,0.49,0.74}
\usepackage[pagebackref,breaklinks,colorlinks,citecolor=cvprblue]{hyperref}
\usepackage{amssymb}
\def\paperID{*****} 
\def\confName{CVPR}
\def\confYear{2024}
\usepackage{setspace}

\title{Integrating Query-aware Segmentation and Cross-Attention for Robust VQA}

\begin{document}

\author{Wonjun Choi$^\dagger$, Sangbeom Lee$^\dagger$, Seungyeon Lee$^\dagger$, Heechul Jung, Dong-Gyu Lee$^\ast$ \\
Department of Artificial Intelligence, Kyungpook National University\\
Daegu, South Korea\\
{\tt\small \{sangju, sb486868, statai3237, heechul, dglee\}@knu.ac.kr}}
\maketitle
\def\thefootnote{$\dagger$}{\footnotetext{Equal Contribution}}
\def\thefootnote{$\ast$}\footnotetext{Corresponding Author}

\begin{abstract}

This paper introduces a method for VizWiz-VQA using LVLM with trainable cross-attention and LoRA finetuning. 
We train the model with the following conditions: 
1) Training with original images. 
2) Training with enhanced images using CLIPSeg to highlight or contrast the original image. 
3) Training with integrating the output features of Vision Transformer (ViT) and CLIPSeg features of the original images.
Then, we ensemble the results based on Levenshtein distance to enhance the prediction of the final answer. In the experiments, we demonstrate and analyze the proposed method's effectiveness.

\end{abstract}
\vspace*{-15pt}
\section{Introduction}

\indent \indent Visual Question Answering (VQA) tasks integrate vision and language, requiring a model to understand and answer questions about a given image. The conventional approaches in VQA have relied on direct mappings between visual content and question-answer pairs, often skipping the nuanced interplay between image-specific details and the context of the questions \cite{lu2023multiscale}. Enhancing the effectiveness of VQA models necessitates attention to global image features and localized, relevant visual cues that directly pertain to the questions asked. The VizWiz-VQA \cite{gurari2018vizwiz} focus on the parts of the image that correspond to the question's intent is crucial. In this paper, we introduce a parameter-efficient method using the Low-rank Adaptation(LoRA) and a cross-attention mechanism between text and visual input based on a large vision language model (LVLM). In addition, we integrate an image segmentation technique, called CLIPSeg \cite{luddecke2022image} to improve visual understanding. We show results using validation/test-dev sets to demonstrate the effectiveness of our method.

\begin{figure}[t]
	\includegraphics[width=0.8\linewidth,height=8cm]{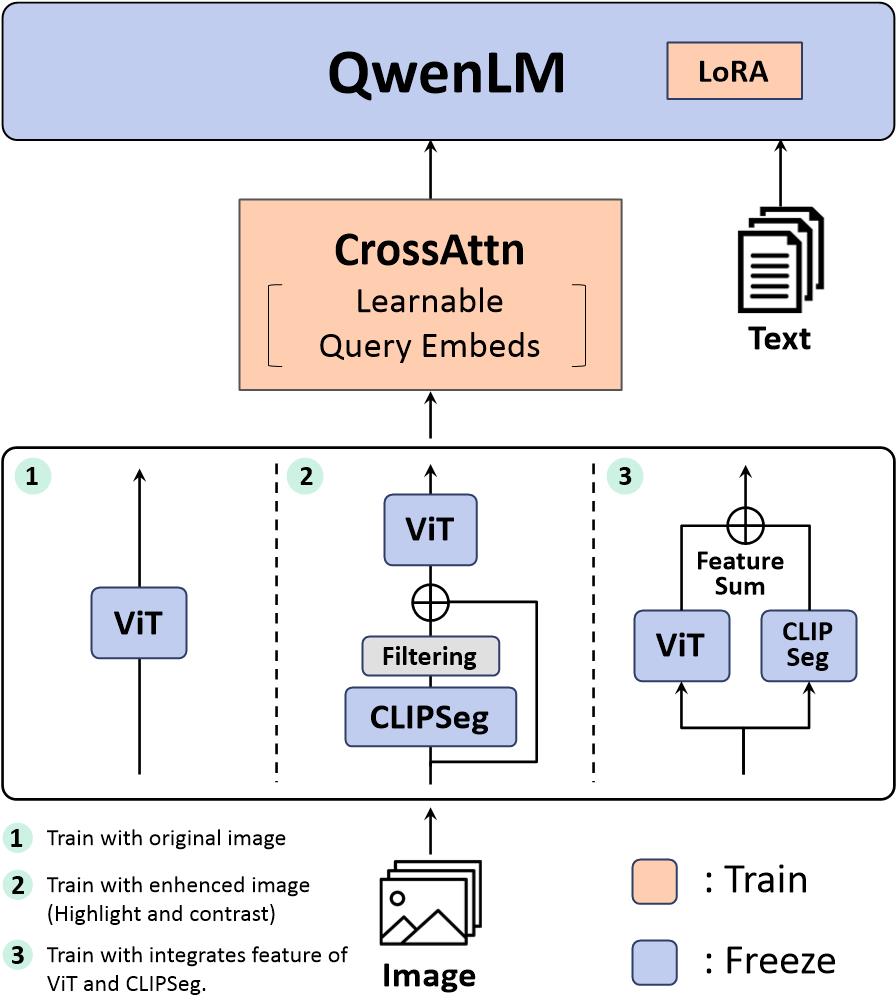}
    \centering
    \caption{Our proposed fine-tuning using LoRA and cross-attention (FT-L/CA) method and three types of ViT configurations.} 
    \label{fig:method} 
\end{figure}

\begin{table*}[t] 
\begin{center}
\begin{tabular}{c|c|c|c|c|c} 
\hline 
Method & yes/no & number & other & unanswerable & overall \\    \hline 

Zero-shot (original image, w/ instruct) & 76.87 & 50.42 & 56.26 & 93.45 & 69.05 \\  
Zero-shot (segment-highlight, w/ instruct) & 64.15 & 42.50 & 49.83 & \textbf{97.23} & 65.59 \\   
Zero-shot (segment-contrast, w/ instruct) & 67.08 & 46.88 & 53.08 & 95.96 & 67.39 \\   \hline  \hline 
FT-L/CA & 81.23 & 52.71 & 58.53 & 93.60 & 70.73 \\   
FT-L/CA (segment-highlight image)& 79.74 & 51.88 & 54.53 & 92.32 & 67.75 \\
FT-L/CA (segment-contrast image)& 79.64 & 48.96 & 57.16 & 93.13 & 69.62 \\
FT-L/CA + Segmentation & \textbf{81.59} & \textbf{56.25} & 58.84 & 91.99 & 70.47 \\  \hline 
Ensemble & 81.38 & 52.71 & \textbf{60.30} & 94.10 & \textbf{72.01} \\  

\hline 
\end{tabular}
\end{center}
\captionsetup{justification=centering}
\caption{Experimental Results on validation set of VizWiz-VQA. FT-L/CA is that fine-tuning \\ with LoRA and cross-attention. The ensemble result is perfomed using fine-tuned models. }
\label{tab:experimental_result_val} 
\end{table*}

\begin{table*}[t] 
\begin{center}
\begin{tabular}{c|c|c|c|c|c} 
\hline 
Method & yes/no & number & other & unanswerable & overall \\    \hline 
 
FT-L/CA & 83.37 & \textbf{55.08} & 58.78 & 94.81 & 69.78  \\   
FT-L/CA (segment-highlight image)& 81.07 & 49.05 & 54.49 & 93.07 & 66.25  \\   
FT-L/CA (segment-contrast image) & 78.76 & 48.10 & 56.63 & 93.45 & 67.66  \\   
FT-L/CA + Segmentation & 81.85 & 50.32 & 56.89 & 87.52 & 66.36 \\   \hline 
Ensemble & \textbf{83.60} & 53.49 & \textbf{60.57} & \textbf{94.88} & \textbf{70.97} \\

\hline  
\end{tabular}
\end{center}
\caption{Experimental Results on test-dev set of VizWiz-VQA.}
\label{tab:experimental_result_test} 
\end{table*}

\section{Method}

\textbf{Trainable cross-attention and LoRA tuning.} We use the Qwen-VL \cite{bai2023qwen} model, which consists of ViT and QwenLM, as a backbone and set the cross-attention module to a learnable state to effectively learn specific parts of the image associated with the keywords in the question. The LoRA adapter \cite{hu2022lora}, which has high parameter efficiency, is applied to fine-tune the QwenLM. In this configuration, only the cross-attention and LoRA application parts are set as trainable parameters, and the rest of the model remains with fixed parameters. \newline 
\textbf{Image segmentation for enhancement.}
We use a query-aware segmentation feature, using the CLIPSeg model \cite{luddecke2022image}, which identifies and separates specific objects within an image based on text queries. CLIPSeg enhances image processing by dynamically isolating related objects and integrating them into the model. We enhance this with segmentation techniques highlighting specific image areas and contrast methods differentiating between masked and unmasked regions, providing more apparent visual distinctions. \newline 
\textbf{Ensemble using Levenshtein distance.} In our proposed method, we employ an ensemble of various predictive models trained to enhance accuracy. For the final answer generation, we utilize a method based on the Levenshtein distance to select the most accurate prediction. This approach leverages the edit distance metric to compare and evaluate the outputs of different models, thereby optimizing the selection of the final answer. 
\section{Experiments}

\noindent \textbf{Experimental results.} Our experiment is conditioned under three conditions, as shown in Figure~\ref{fig:method}.
(1) Input the original image into Qwen-VL's ViT(FT-L/CA).
(2) Input the original image with two variants segmentation (highlight, contrast). The segment-highlight adds the segmentation output to the original image, and segment-contrast adjusts the brightness of the original image other than segmentation output. (3) Integrate ViT features and CLIPSeg features in the model.
We show the results of applying segmentation as examples in the appendix \ref{appendix}.
In our experiments, we perform experiments on zero-shot and fine-tuning. In the zero-shot setting, we confirmed without instructions, performance deteriorates significantly, thus we add and use instructions\footnote{\textbf{Instructions: }`` The correct answer type is one of [`number', `words', `yes', `no']. If it is impossible to answer an image-related question or there is no existing information, please reply as `unanswerable'. Question: ''}. 
As shown in Table~\ref{tab:experimental_result_val} and Table~\ref{tab:experimental_result_test}, integrating feature information directly in the model proves more effective than applying CLIPSeg output to the input image and transforming it, and particularly, the ensemble model yields the highest performance.
The query-aware CLIPSeg actively improves correct answer rates across multiple categories.
\section{Conclusion}

\indent \indent In this paper, we present a method using LVLM with trainable cross-attention and LoRA tuning. We also use query-aware CLIPSeg to enhance focus on relevant image details. Our method ranked 3rd (team: KNU-HomerunBall) in the VizWiz challenge public leaderboard, demonstrating improved results through further experiments and ensembles.
\section*{Acknowledgements}
This work was supported by the National Research Foundation of Korea (NRF) grant funded by the Korean Government (MSIT) (No. 2021R1C1C1012590), (No. 2022R1A4A1023248), and (No. 2022R1A5A7026673).

{
    \small
    \bibliographystyle{ieeenat_fullname}
    \bibliography{main}
    
}

\newpage 
\appendix 
\onecolumn
\section{Appendix} \label{appendix}

\setcounter{figure}{0}
\renewcommand{\thefigure}{A\arabic{figure}}

\begin{figure*}[h]
	\includegraphics[width=0.8\linewidth]{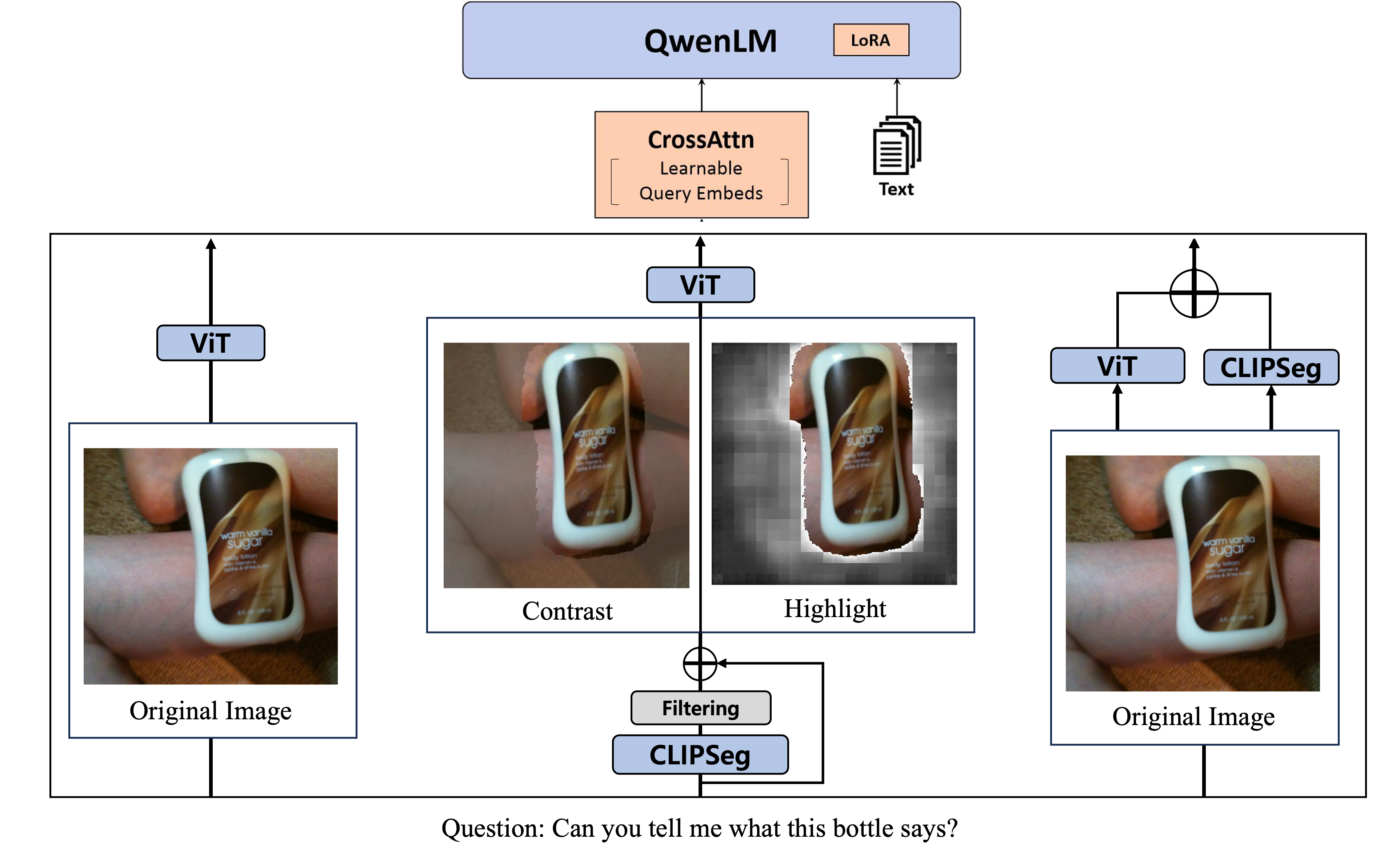}
    \centering
    \caption{Example of an original input image (bottle) and the result of applying CLIPSeg.} 
    \label{fig:example1} 
\end{figure*}

\begin{figure*}[h]
	\includegraphics[width=0.8\linewidth]{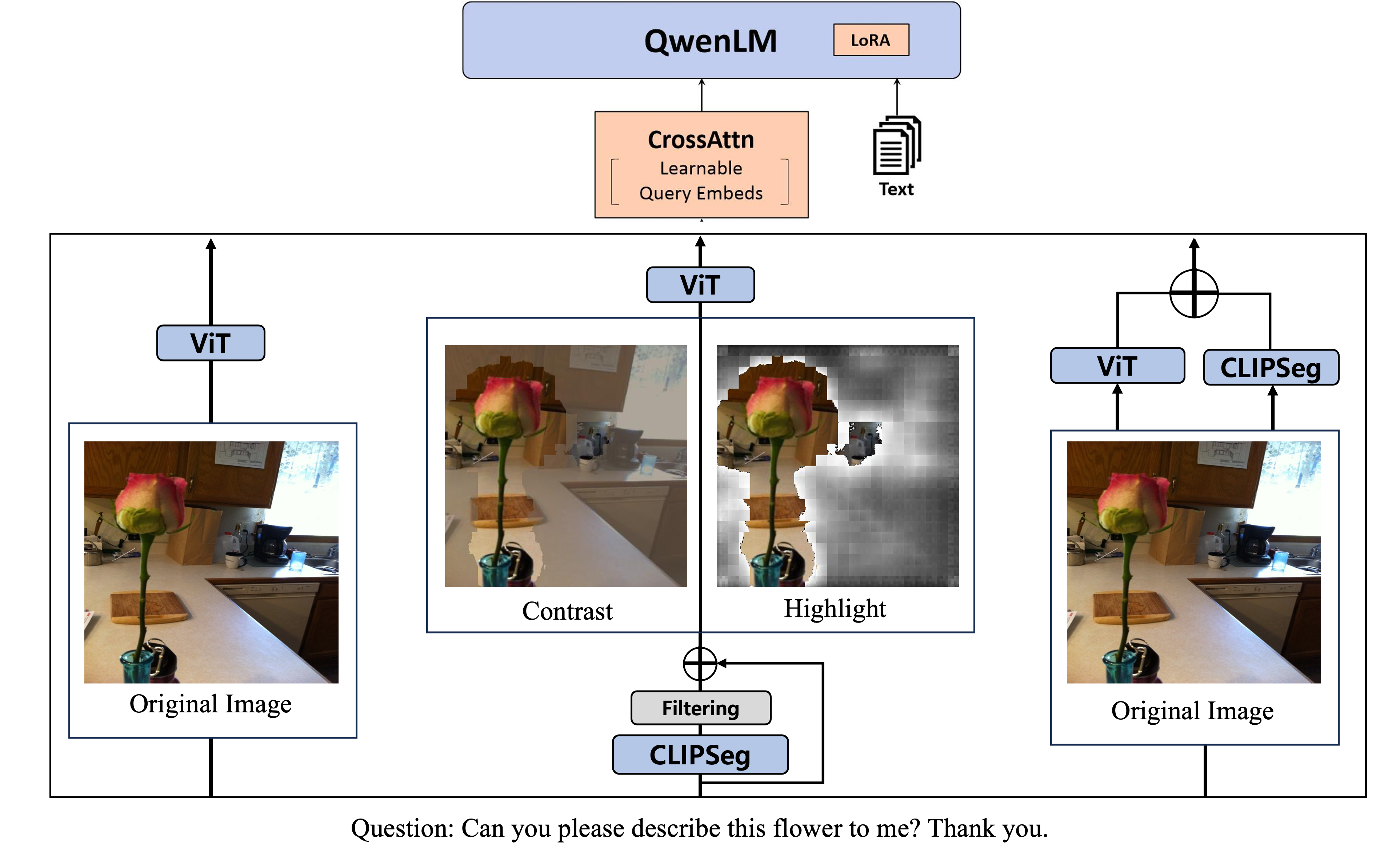}
    \centering
    \caption{Example of an original input image (flower) and the result of applying CLIPSeg.} 
    \label{fig:example2} 
\end{figure*}

\end{document}